\documentclass{article}

\usepackage{microtype}
\usepackage{graphicx}
\usepackage{subfigure}
\usepackage{booktabs} 
\usepackage{multirow}
\usepackage{tikz}
\usetikzlibrary{shapes}
\usepackage{pgfplots}
\usepackage{hyperref}

\usepackage{algorithmic}

\usepackage[accepted]{icml2023}

\usepackage{amsmath}
\usepackage{amssymb}
\usepackage{mathtools}
\usepackage{amsthm}

\usepackage[capitalize,noabbrev]{cleveref}

\theoremstyle{plain}
\newtheorem{theorem}{Theorem}[section]
\newtheorem{proposition}[theorem]{Proposition}
\newtheorem{lemma}[theorem]{Lemma}
\newtheorem{corollary}[theorem]{Corollary}
\theoremstyle{definition}
\newtheorem{definition}[theorem]{Definition}
\newtheorem{assumption}[theorem]{Assumption}
\theoremstyle{remark}
\newtheorem{remark}[theorem]{Remark}

\pgfplotsset{compat=1.18}

\usepackage{environ}
\NewEnviron{draft}{
   {\color{gray}\BODY}
  }

\usepackage[textsize=tiny]{todonotes}

\icmltitlerunning{Information-Theoretic Approach to Keyword Extraction}

\newcommand{\Pmodel}{P_{\mathsf{model}} }

\definecolor{HCcolor}{HTML}{d1615d}
\definecolor{LRcolor}{HTML}{5778a4}

\newcommand{\note}[1]{{\sf\textcolor{red}{[#1]}}}

\begin{document}
\twocolumn[
\icmltitle{EntropyRank: Unsupervised Keyphrase Extraction via Side-Information Optimization for Language Model-based Text Compression}




\begin{icmlauthorlist}
\icmlauthor{Alexander Tsvetkov}{runics,microsoft}
\icmlauthor{Alon Kipnis}{runics}
\end{icmlauthorlist}

\icmlaffiliation{runics}{School of Computer Science, Reichman University, Herzliya, Israel}

\icmlaffiliation{microsoft}{Microsoft, Herzliya, Israel}
\icmlcorrespondingauthor{Alexander Tsvetkov}{alexander.tsvetkov@post.runi.ac.il}
\icmlcorrespondingauthor{Alon Kipnis}{alon.kipnis@runi.ac.il}

\icmlkeywords{keyword extraction, logloss, entropy, NLP, lossy compression, summarization}

\vskip 0.3in
]



\printAffiliationsAndNotice{\icmlEqualContribution} 

\begin{abstract}
We propose an unsupervised method to extract keywords and keyphrases from texts based on a pre-trained language model (LM) and Shannon's information maximization. Specifically, our method extracts phrases having the highest conditional entropy under the LM. The resulting set of keyphrases turns out to solve a relevant information-theoretic problem: if provided as side information, it leads to the expected minimal binary code length in compressing the text using the LM and an entropy encoder. Alternately, the resulting set is an approximation via a causal LM to the set of phrases that minimize the entropy of the text when conditioned upon it. Empirically, the method provides results comparable to the most commonly used methods in various keyphrase extraction benchmark challenges.
\end{abstract}

\section{Introduction}
\label{sec:intro}

\subsection{Motivation}

Keyphrase extraction can be described as an information distillation process of a document into a series of words. These words are later used as a proxy for document representation, to be utilized in various downstream tasks such as extractive summarization, information retrieval, clustering, document categorization, and query expansion \cite{hasan2014automatic, medelyan2008domain}. 
Although the problem involves a task that is native to information theory, i.e., extracting information subject to a constraint, to the best of our knowledge, non of the existing methods directly optimize Shannon's information or strive to solve an information transmission problem. The purpose of the current work is to suggest a method that stems from such information-theoretic principles and to demonstrate that it performs comparable to the state-of-the-art methods. 


%


\subsection{Contribution}
We present a novel method of unsupervised keyphrase extraction denoted EntropyRank that strives to minimize Shannon's entropy of text given the keyphrases. The conditional entropy is evaluated with respect to a pre-trained LM, typically a large LM based on transformer deep neural networks \cite{vaswani2017attention}. 
The resulting set of keyphrases has a relevant operational interpretation: the set that provides the maximal reduction in the expected binary code length (bits) when compressing the text using the LM and an entropy encoder, while the keyphrases and their locations are provided as side information. Works utilizing this form of compression but without side information are known to achieve state-of-the-art results on lossless text compression \cite{izacardlossless,mahoney2011large}. The extraction principle of EntropyRank is reminiscent of lossless compression methods that provide the most difficult parts to predict as side information 
\cite{caire2003lossless}. 

While our method is derived directly from information theoretic principles, it appears to perform well empirically, attaining results comparable to the most commonly used method over a series of benchmark tasks; see the report in Section~\ref{sec:results}. 

\subsection{Background}
%
Keyphrase extraction can be supervised or unsupervised, with the former requiring labeled training data and the latter being more domain-independent \cite{sahrawat2020keyphrase}. In many situations, manual labeling is impractical or unavailable due to domain adaptation challenges hence unsupervised keyphrase extraction is the only viable option. 

Unsupervised keyphrase extraction methods can be categorized into four groups based on the features they use to rank candidates: statistical, embedding-based, graph-based, and generative. Statistical methods such as RAKE and YAKE use a pre-trained model of combining features such as TF-IDF, relative position, and co-occurrence \cite{campos2018yake}. 
Embedding-based methods such as KeyBert and PatternRank use word or sentence embeddings to measure the relevance of text chunks to the document \cite{bennani2018simple, grootendorst2020keybert, schopf2022patternrank}. Graph-based methods such as TextRank construct a co-occurrence graph of words or phrases and apply centrality measures to score them \cite{mihalcea2004textrank, rose2010automatic}. Some methods also combine textual or semantic features with graph or embedding features to form hybrid models\cite{mahata2018key2vec, mahata2018theme}. 
The newest generative methods use instruction-tuned LMs \cite{ouyang2022training} directly with a prefixed context of the text and an instruction in the form of a contextualized prompt to generate keywords directly from the text. 
However, most of these approaches have some inherent drawbacks. Specifically, statistical and graph-based methods rely on local corpus features, hence they disregard natural language regularities and typically require some hyper-parameter tuning. 
Semantic and embedding methods tend to struggle with phrases that do not match the document's context without proper tuning. Generative models, on the other hand, can produce unreliable and unpredictable results due to biases and hallucinations \cite{ji2023survey}, which hinder their use in practice. In contrast, EntropyRank incorporates language regularities and semantics directly from the LM, suggesting that it works well whenever the LM reasonably predicts tokens under a cross-entropy (log) loss, the typical training objective of modern LMs.

\subsection{Organization}
The rest of this paper is organized as follows. In Section~\ref{sec:method} we describe the method. In Section~\ref{sec:IT} we analyze the method under the lens of source coding in information theory. In Section~\ref{sec:results} we report on empirical results. Concluding remarks are provided in Section~\ref{sec:conclusions}.

%
\section{Method Description}
\label{sec:method}
Let $\Pmodel$ be a causal LM, i.e., a set of conditional probability distributions over sequences of tokens $w_{1:n}=(w_1,\ldots,w_n)$ of the form 
\[
\Pmodel(\cdot|w_{1:i-1}) = \Pr[W_i|W_{1:i-1}=w_{1:i-1}],\quad i\leq n.
\]
Above and throughout we use the notation  $u_{1:0} := \emptyset$ for any sequence $u$. By extension, $\Pmodel$ also provides conditional probabilities of the form $\Pr[X_i|X_{1:i-1}=x_{1:i-1}]$ where $x_{1:n} = (x_1,\ldots,x_n)$ is a sequence of text phrases and each phrase is a sequence of tokens. Specifically, if phrase $x_i$ consists of tokens $(w_{i,1},\ldots,w_{i,n_i})$, then the probability of $x_i$ is
\[
\Pmodel(x_i|x_{1:i-1}) = \prod_{j=1}^{n_i} \Pmodel(w_{i,j} |x_{1:i-1}, w_{i,1:j-1}).
\]
Our method first segments the document into phrases, for example using noun phrases or stop words as in \cite{schopf2022patternrank, rose2010automatic}. Given the segmented document $x_{1:n} = (x_1,\ldots, x_n)$, we refer to
    \begin{align}
        \label{eq:H_i_def}
        H_i :=  H(X_i|X_{1:i-1}=x_{1:i-1}) = H(\Pmodel(\cdot|x_{1:i-1})), 
    \end{align}
    as the entropy of the $i$-th phrase under the LM. Here $\Pmodel(\cdot|x_{1:i-1})$ is the distribution of the $i$-th phrase in the document given the previous $i-1$ phrases as provided by the LM and $H_i$ is Shannon's entropy of this distribution \cite{cover2012elements}. Note that $H_i$ is a function of the phrases preceding $x_i$ but not of $x_i$ itself. 

    Our method outputs a set of phrases $\{X_j,\,j \in J^*\}$, $J^*\subset\{1,\ldots,n\}$, that maximizes the sum of phrase entropies subject to the cardinality constraint of at most $k$ elements. Namely, $J^*$ maximizes
    \begin{align}
    \label{eq:optimization}
    \sum_{j\in J} H_j,\quad \text{subject to}\quad |J^*|\leq k; 
    \end{align}
    the entire procedure is summarized in Algorithm~\ref{alg:1}. 
    Another useful practice is to report the smallest set of keyphrases such that the sum in \eqref{eq:optimization} exceeds some specified bit threshold. An operational interpretation of the phrase entropies and this bit threshold is given next. 

\begin{algorithm}[tb]
   \caption{EntropyRank
   \label{alg:1}
   }
   \label{alg:example}
   
\begin{algorithmic}
   \STATE {\bfseries Input:} Text document $D$, number of keyphrases $k$, language model $\Pmodel$
   \STATE Segment text to phrases $x_{1:n}=(x_1,\ldots,x_n)$
   \FOR{$i=1$ {\bfseries to} $n$}
   \STATE $H_i \leftarrow H(\Pmodel(\cdot | x_{1:i-1}))$
   \ENDFOR
   \STATE $J^* = \underset{{J\,:\,|J|\leq k}}{\arg\max} \sum_{j \in J} H_i$
   \STATE {\bfseries return} $\{X_j \}_{j \in J^*}$
\end{algorithmic}
\end{algorithm}



\section{Information Theoretic Analysis}
\label{sec:IT}
We provide two viewpoints to motivate EntropyRank. 

\subsection{Lossless text compression with side information}
Consider the problem of compressing the text using a binary code when a set of phrases indexed by $J$ is provided as side information while the cardinality of $J$ is restricted. We can interpret the entropy $H_i$ of \eqref{eq:H_i_def} as the amount of information keyphrase $x_i$ provides on the text in the following sense. It is a lower bound on the expected reduction in the number of bits needed to encode the text using the LM $\Pmodel$ and an entropy encoder when $x_i$ and its location $i$ are provided as side information (regardless of the distribution of the text). To better explain what we mean by this kind of encoding, suppose that we encode $x_{1:n}$ when $(x_m, m)$, $2 < m \leq n$ is provided as side information and using an arithmetic encoder as the entropy encoder  \cite{langdon1984introduction}. Starting with $x_1$, we find a partition of the interval $[a_0,b_0)=[0,1)$ according to the distribution $\Pmodel(\cdot|\emptyset)$ in a pre-determined order; we denote by $[a_1,b_1)$ the interval corresponding to $x_1$ in this partition. Next, we partition $[a_1,b_1)$ according to $\Pmodel(\cdot|x_1)$ in the same order and denote by $[a_2,b_2)$ the interval corresponding to $x_2$ in this partition. The situation continues until we reach $x_m$, in which case we add $x_m$ as a context to the LM but otherwise ignore it and move to partition $[a_{m-1},b_{m-1})$ according to $\Pmodel(\cdot|x_{1:m})$. The resulting encoded representation of the text is the shortest binary representation falling within the interval corresponding to $x_n$ at the last step with the leading zero removed. This encoding process is clearly reversible given $\Pmodel$ and $(x_m, m)$. The extension to more than one phrase provided as side information is straightforward. This form of encoding but without incorporating side information is used in \cite{izacardlossless, liu2019decmac, goyal2021dzip}. 
\begin{figure}
    \centering
    \begin{tikzpicture}
 \begin{axis}[
    width=8cm,
    height=4cm,
    legend style={at={(0.9,1)},
      anchor=north east, legend columns=1},
    ylabel={\small $H({\bar{J}^*})$ [bits/word]},
    xlabel={\small \# of keyphrases},
    xtick={0,1,2,3,4,5,6,7,8,9,10,11,12,13,14,15},
    xticklabels={0,,2,,4,,6,,8,,10,,12,,14,,},
    ymin=3,
    xmin=0,
    xmax=15,
    ymax=5.5,
    ]
\addplot [only marks, color=red] table [col sep=comma] {average_entropy_per_word_in_abstract_without_key_phrases_at_k_total_words_normalization.csv};


\end{axis}
\end{tikzpicture}
      \vspace{-1\baselineskip}
    \caption{
    Expected remaining normalized text entropy $H({\bar{J}^*})$ versus the number of keyphrases. 
    The remaining entropy is the expected number of bits needed to encode the text via a LM of matching distribution and an entropy encoder when the keyphrases are provided as side information. 
    }
    \label{fig:entropy_per_no_keyphrases}
\end{figure}
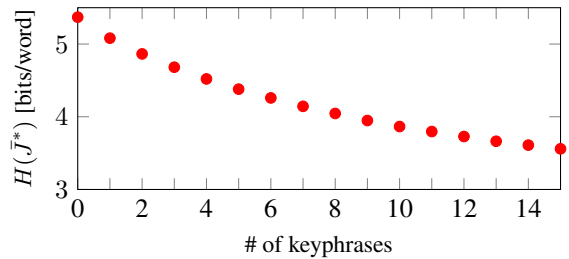
Our method ranks each phrase according to the entropy $H_i$ of \eqref{eq:H_i_def} associated with its location in the text. By design, the highest-ranked keyphrase provides more information (in the sense of expected code length reduction) on the text than the second-highest, and so forth. 
A lower bound on the expected code length is the remaining text entropy provided by the sum $H({\bar{J}^*}):=\sum_{j \notin J^*} H_i$. This lower bound is asymptotically attained when the LM exactly describes the text's distribution in the sense of vanishing relative entropy rate \cite{gray2011entropy}. 
This can be seen empirically in Figure~\ref{fig:entropy_per_no_keyphrases}, showing the average entropy of keyphrases based on their rank over the Inspec dataset \cite{hulth2003improved}. 

\subsection{Approximating the Information Maximizing Set}
EntropyRank also arises as a tractable approximation to an optimal information-theoretic solution of the keyphrases extraction problem. In order to formulate this problem, denote by $X_{1:n} = (X_1,\ldots,X_n)$ the sequence of phrases constituting a document, viewed as random variables over a dictionary. Let $J \subset \{1,\ldots,n\}$ be a set of indexes of phrases in this document. 
We seek a set of keyphrases indexed by $J^\dagger$ that captures most of the information as measured by Shannon's entropy $H(X_{1:n})$. Namely, with $\bar{J} = \{1,\ldots,n\} \setminus J$, $J^\dagger$ minimizes $H(X_{\bar{J}}|X_J)$ in the decomposition
\begin{align}
    \label{eq:H_decomp}
H(X_{1:n}) = H(X_J) + H(X_{\bar{J}}| X_J)
\end{align} 
subject to the cardinality constraint $|J|\leq k$. Since the mutual information decomposes as
\begin{align*}
I(X_{1:n}; X_J) & = H(X_{1:n}) - H(X_{1:n}|X_J) \\
& = H(X_{1:n}) - H(X_{\bar{J}}|X_J),
\end{align*}
we can also think of $J^\dagger$ as the maximizer of the mutual information between the set of keyphrases and the entire text subject to the cardinality constraint.

It is usually intractable to minimize $H(X_{\bar{J}}| X_J)$ directly and evaluate $J^\dagger$ for large texts due to the large search space and the need to evaluate non-causal conditional probability expressions\footnote{We only have an interface to evaluate causal conditional probabilities of the form $\Pr[X_i|X_{1:i-1}=x_{1:i-1}]$, hence we must marginalize over the dictionary to evaluate non-causal probabilities like $\Pr[X_2|X_1=x_1,X_3=x_3]$. Each marginalization involves several thousands of LM evaluations hence the entire search quickly becomes impractical.}. We turn to seek an approximation to $J^{\dagger}$. We decompose the entropy of the text as
\begin{align}
    \label{eq:H_decomp_alt}
H(X_{1:n}) = \sum_{i \in J} H(X_i |X_{1:i-1})  + \sum_{i \in \bar{J}} H(X_i |X_{1:i-1}) 
\end{align}
and look for a set $J^*$ that maximizes $\sum_{i \in J} H_i$, i.e. the observed version of the first sum in \eqref{eq:H_decomp_alt}, subject to the cardinality constraint. Our method provides the set $J^*$ under the assumption that the conditional distribution of the text is provided by the LM. In the case of a distributional mismatch between the LM and the text, an analogous logic applies when replacing entropy with cross-entropy \cite{cover2012elements}.

\begin{table*}[t]
\label{keyphrase-table}
\vskip 0.15in
\begin{center}
\begin{small}
\begin{sc}
\begin{tabular}{l|l|cccc|cccc|cccc}
\toprule
\multirow{ 1}{*}{} & \multirow{ 2}{*}{METHOD} & \multicolumn{4}{c|}{@5 keyphrases} & \multicolumn{4}{c|}{@10 keyphrases} \\
\cr{}
 &  & P & R & F1 & RO1 & P & R & F1 & RO1 \\
\midrule
\multirow{ 5}{*}{\rotatebox[origin=c]{90}{Inspec}} 
 & PatternRank & \textbf{32.9} & \textbf{30.99} & \textbf{29.42} & \textbf{44.51} & \textbf{28.5} & \textbf{49.7} &\textbf{33.85} & \textbf{48.71} \\
& EntropyRank & \underline{32.21} & \underline{29.18} & \underline{28.26} & \underline{43.8} & \underline{27.47} & \underline{47.12} & \underline{32.39} & \underline{48.15}   \\
 & RAKE & 21.34 & 20.6 & 19.32 & 37.39 & 22.24 & 39.71 & 26.63 & 43.29 \\
 & YAKE & 17.33 & 17.02 & 15.74 & 33.3 & 14.34 & 27.4 & 17.46 & 30.04 \\
 & TextRank & 30.45 & 27.83 & 26.86 & 39.61 & 25.51 & 44.52 & 30.24 & 43.8 \\
\midrule
\multirow{ 5}{*}{\rotatebox[origin=c]{90}{SE 2010}}
 & PatternRank & 7.95 & 4.73 & 5.75 & \textbf{23.41} & 6.8 & 7.83 & 7.06 & \textbf{21.82} \\
& EntropyRank & \underline{4.92} &  \underline{2.56} & \underline{3.31} & \underline{15.31} &  \underline{5.53} & \underline{5.93} & \underline{5.58} & \underline{18.98} \\
 & RAKE & 0.08 & 0.05 & 0.06 & 5.18 & 0.04 & 0.05 & 0.04 & 9.17 \\
 & YAKE & \textbf{11.72} & \textbf{6.31} & \textbf{7.98} & 16.97 & \textbf{10.45} & \textbf{11.06} & \textbf{10.46} & 20.59 \\
 & TextRank & 4.84 & 2.63 & 3.3 & 14.88 & 4.1 & 4.44 & 4.12 & 14.81 \\
\midrule
\multirow{ 5}{*}{\rotatebox[origin=c]{90}{SE 2017}} 
 & PatternRank & \textbf{35.52} & \textbf{16.24} & \textbf{21.43} & \textbf{29.08} & \textbf{32.04} & \textbf{28.5} & \textbf{28.87} & \textbf{42.9}  \\
& EntropyRank & \underline{28.36} & \underline{12.88} & \underline{17.0} & \underline{26.15} & \underline{25.82} & \underline{23.09} & \underline{23.3} & \underline{39.72} \\
 & RAKE & 18.0 & 8.52 & 11.12 & 25.01 & 20.64 & 18.78 & 18.89 & 39.53 \\
 & YAKE & 18.16 & 8.2 & 10.84 & 19.85 & 17.86 & 16.12 & 16.17 & 28.58 \\
 & TextRank & 25.68 & 11.73 & 15.48 & 25.6 & 24.2 & 21.59 & 21.88 & 37.59 \\
\bottomrule
\end{tabular}
\end{sc}
\caption{Performance evaluation of keyphrase extraction models on benchmark datasets. \textbf{Bolded} values indicate highest score, \underline{underlined} values indicate our method.}
\label{table:benchmark_eval}
\end{small}
\end{center}
\vskip -0.1in
\end{table*}

\section{Empirical Results}
\label{sec:results}

\subsection{Implementation}
We use GPT-Neo 1.7B \cite{gpt_neo}, a pre trained LM trained on the PILE dataset \cite{gao2020pile}, to estimate the natural language distribution of the text. We segment the text into noun phrases that match the parts of speech tag patterns
\textless J.*\textgreater *\textless N.*\textgreater+, capturing zero or more adjectives followed by one or more nouns. We rank these segments by the sum of the entropy of their words and extract the top $k$ candidates with the highest entropy as keyphrases.

\subsection{Datasets}
We evaluated our method on three common datasets with expert annotations, which are often used to evaluate keyphrase extraction methods in the literature.

Inspec \cite{hulth2003improved}- abstracts of 2,000 English scientific papers from the Inspec database.

SE-2010 \cite{kim-etal-2010-semeval}- full scientific articles that are obtained from the ACM Digital Library.

SE-2017 \cite{augenstein2017semeval}- abstracts of 500 English scientific papers from the ScienceDirect publications.
\subsection{Baseline Methods}

We compared our method to popular baseline methods:

\textbf{PatternRank} \cite{schopf2022patternrank} - an extension of KeyBERT which extracts the noun phrases with the highest document similarity.

\textbf{RAKE} \cite{rose2010automatic} - extracts phrases based on delimiters(stopwords, punctuation) and co-occurrences scoring.

\textbf{YAKE} \cite{campos2018yake} - based on statistical features such as term frequency, and position.

\textbf{TextRank} \cite{mihalcea2004textrank} -  applies a graph-based ranking algorithm to words and phrases.

\subsection{Evaluation Metrics}
To assess the quality of our key phrase extraction method, we used classification and summarization metrics. The former included recall, precision, and f1 scores at different $k$ values, measuring the agreement with the ground truth labels. The latter was ROUGE1\cite{lin-2004-rouge}, which calculates the single word overlap between the concatenated key phrases and the gold key phrases, reflecting the information distillation aspect of the task.

\subsection{Discussion}
The benchmark results on Table \ref{table:benchmark_eval} show that EntropyRank performs well on short text datasets, such as SE2017 and INSPEC, where it achieves similar results to PatternRank and surpasses all the other methods. However, it struggles with long texts, such as SE2010, possibly due to the low rank our method gives to phrases with many occurrences which in long texts are more likely to be keyphrases. This limitation is easy to resolve in practice by using a simple term frequency-based extractor in parallel to EntropyRank. A comparison of the extracted key phrases by EntropyRank and PatternRank on INSPEC reveals a low Jacard similarity score of 0.21, indicating that they produce different and complementary results. Thus, EntropyRank has shown to be a suitable keyphrase extraction method for short texts and can enhance other methods as a complementary approach.
\section{Conclusions}
We presented EntropyRank, a novel unsupervised method for keyphrase extraction based on the information-theoretic principles of conditional entropy minimization under a pre-trained language model. The method is simple and very direct to apply. Nevertheless, empirical results demonstrate that our method is comparable to state-of-the-art methods on several benchmark challenges. 

In future work, we plan to explore the connection between our method and task-oriented lossy compression. For example, by evaluating the impact of our keyphrases on downstream tasks such as IR, clustering, or categorization.
\label{sec:conclusions}

\newpage
\bibliography{main}
\bibliographystyle{icml2023}

\end{document}